\documentclass[12pt]{iopart}
\usepackage{iopams} 
\usepackage[backref]{hyperref}
\usepackage{cite}
\usepackage{color}
\usepackage{graphicx}
\usepackage{booktabs}
\usepackage{multirow}
\usepackage{threeparttable}
\usepackage{makecell}

\graphicspath{{./fig/}}

\begin{document}

\title[NeuSort: an automatic adaptive spike sorting approach]{NeuSort: an automatic adaptive spike sorting approach with neuromorphic models}

\author{
  Hang Yu$^{1,2}$,
  Yu Qi$^{1,3,4}$\footnote{Corresponding authors: Yu Qi and Gang Pan.}, and
  Gang Pan$^{1,2}$\footnotemark[1]}

\address{$^1$ State Key Lab of Brain-Machine Intelligence, Hangzhou, China}
\address{$^2$ College of Computer Science and Technology, Zhejiang University, Hangzhou, China}
\address{$^3$ Affiliated Mental Health Center \& Hangzhou Seventh People's Hospital, Hangzhou, China}
\address{$^4$ MOE Frontier Science Center for Brain Science and Brain-machine Integration, Zhejiang University School of Medicine, Hangzhou, China}
\ead{\mailto{yuh@zju.edu.cn}, \mailto{qiyu@zju.edu.cn}, \mailto{gpan@zju.edu.cn}}

\begin{abstract}
\textit{Objective.} 
Spike sorting, a critical step in neural data processing, aims to classify spiking events from single electrode recordings based on different waveforms. 
This study aims to develop a novel online spike sorter, NeuSort, using neuromorphic models, with the ability to adaptively adjust to changes in neural signals, including waveform deformations and the appearance of new neurons.
\textit{Approach.} 
NeuSort leverages a neuromorphic model to emulate template-matching processes. 
This model incorporates plasticity learning mechanisms inspired by biological neural systems, facilitating real-time adjustments to online parameters.
\textit{Results.} 
Experimental findings demonstrate NeuSort's ability to track neuron activities amidst waveform deformations and identify new neurons in real-time. 
NeuSort excels in handling non-stationary neural signals, significantly enhancing its applicability for long-term spike sorting tasks.
Moreover, its implementation on neuromorphic chips guarantees ultra-low energy consumption during computation.
\textit{Significance.} 
NeuSort caters to the demand for real-time spike sorting in brain-machine interfaces through a neuromorphic approach.
Its unsupervised, automated spike sorting process makes it a plug-and-play solution for online spike sorting.
\end{abstract}

\submitto{\JNE}

\maketitle

\section{Introduction}
Analysis of individual neuron activities plays a crucial role in neuroscience research \cite{carlson2019continuing, souza2019spike, Yger2018}.
Typically, extracellular recordings from a single electrode contain spike firing activities from several nearby neurons, and different neurons generate action potentials with diverse extracellular waveforms \cite{Huang2021}.
Spike sorting, a technique commonly used, aims to classify the activities of neurons based on the similarity of waveform shapes, allowing for further analysis in neuroscience.

In recent decades, the advancement of large-scale neural recordings has presented new demands and challenges for spike sorting techniques \cite{bod2022end}. 
This trend has spurred the development of software and tools focused on automated spike sorting. 
Traditionally, spike sorting is performed offline, where the sorting process occurs after all the data has been collected.
However, with the emergence of closed-loop brain-computer interfaces and \textit{in vivo} signal analysis, the need for online spike sorting has become crucial \cite{wu2013convergence, qi2022dynamic}.
In online spike sorting, the spike events need to be classified immediately after the signal is recorded. 
Unfortunately, most offline spike sorters, which rely on fixed models established during a setup or training phase, tend to exhibit lower performance in an online setting.
This is because the properties of neurons can change over time, resulting in a decrease in the performance of a fixed model.
As demonstrated in previous research \cite{doi:10.1126/science.abf4588}, \textit{in vivo} recordings often experience cell drift or rotation from their initial positions, attributed to pressure release after probe insertion.
The instability of signal recording \cite{Quiroga2004} can lead to waveform distortions, the disappearance of neurons, and the appearance of new neurons, as the position of the recording probe may shift during the recording process.
In this study, we simplify waveform deformations as a global scaling problem, as discussed in the work by \cite{kumar2022tracking}.

The challenge of addressing waveform deformations and detecting newly emerging neurons in online spike sorting processes is a significant hurdle.
Template-matching-based approaches commonly employ a strategy of correlating the incoming signal with pre-existing spike waveforms.
Previous studies \cite{kyung2003, Brychta2007, Chaure2018} have employed discrete wavelet transform (DWT) to establish correlations between the signal and waveform libraries.
A notable work, SpykingCircus \cite{Yger2018}, integrated density-based clustering and template matching into iterative steps, where the raw data was masked with isolated waveforms from previous cycles. 
Alternatively, non-rigid spatial registration of extracellular signals can be utilized to estimate deviations.
The Neuropixels system \cite{doi:10.1126/science.abf4588} employed an average template of the mean activity histograms computed by record chunk to estimate the offset along the depth dimension of the electrodes.
In a subsequent study, Boussard et al.\cite{boussard2021three} extended this approach to account for probe drifts in three dimensions.
However, template-based methods require manual interaction for template design, and the user must be aware of the spike waveforms of interest in advance. 
Additionally, these approaches often struggle with identifying newly emerging neurons in online processing settings. 

Recently, there has been a growing trend in the use of deep learning (DL) approaches for spike sorting \cite{Rokai_2021}. 
Some studies \cite{radmanesh2022online, EOM2021131, Rokai_2021} employ autoencoders and their variants as data-driven feature reduction techniques to map the original waveforms into a lower-dimensional space, thereby mitigating the impact of waveform deformations on the raw signals.
Another strategy in employing DL models is the use of end-to-end solutions, which aim to take the raw (or filtered) traces as input and directly output the spike trains of the sorted neurons.
Racz et al. \cite{Racz_2020} implemented separate networks for spike detection and spike sorting using a combination of Convolutional Neural Networks (CNN) and Long Short-Term Memory (LSTM). 
Their approach was applied to a 128-channel high-density device. 
Similarly, Li et al. \cite{li2020accurate} developed a similar approach focused on recordings from single channels.
However, it should be noted that these DL methods require higher computational resources compared to traditional approaches. 
Additionally, a substantial amount of training data is required for cold-start initialization. 
These factors should be taken into consideration when adopting DL-based spike sorting techniques.

Neuromorphic models have emerged as a promising approach for spike sorting in recent studies, showcasing the inherent advantages of efficient online learning \cite{ijcai2018p221, imam_rapid_2020, Chakraborty2021, Parameshwara2021, rokai2023edge}. 
Unlike traditional machine learning models that rely on static mathematical models, neuromorphic models, specifically spiking neural networks, possess natural online adaptive capabilities through plasticity-driven updating rules.
These rules allow the model parameters to be adjusted in an unsupervised manner.
Consequently, neuromorphic models offer a natural solution for online spike sorting by effectively handling signal variations and the emergence of new neurons.

In this paper, we introduce NeuSort, an automatic spike sorting method that leverages the power of neuromorphic computing. 
The key improvement of NeuSort lies in its ability to address the continuous and delayed nature of waveform changes. 
By maintaining a constantly updated waveform library, NeuSort enables adaptive online tracking of neurons, ensuring accurate spike sorting in real-time.
Through the utilization of plasticity learning rules inherent in neuromorphic models, NeuSort undergoes online updates, allowing for sequential processing of spike signals and rapid convergence to a spike sorter within a few minutes. 
This feature makes NeuSort a plug-and-play module for online spike sorting, significantly reducing the need for human intervention.
Furthermore, the adaptive nature of the neuromorphic model in NeuSort enables it to automatically adjust to changes in spike waveforms. 
This adaptability plays a crucial role in effectively tracing waveform deformations in neural signals, making NeuSort particularly well-suited for long-term recordings where waveform changes may occur. 
By enhancing the robustness against non-stationary neural signals with spike waveform deformation, NeuSort provides stable and reliable online spike sorting over extended periods.
The proposed neuromorphic spike sorter demonstrates great potential, especially in ultra-low-cost spike sorting applications. 
This is particularly relevant in the context of neuromorphic chips for implantable Brain-Computer Interface (BCI) devices, where efficient and accurate online spike sorting is of utmost importance.

The organization of this paper is as follows. 
In Section~\ref{section_methods}, we describe the structure of our proposed method. 
The experimental results are presented and discussed in Section~\ref{section_experimental_results}. 
Section~\ref{section_discussion} provides an in-depth discussion of our findings, 
while Section~\ref{section_conclusion} summarizes the conclusions drawn from our work.

\begin{figure*}[htbp]
	\begin{center}
		\includegraphics[width=\textwidth]{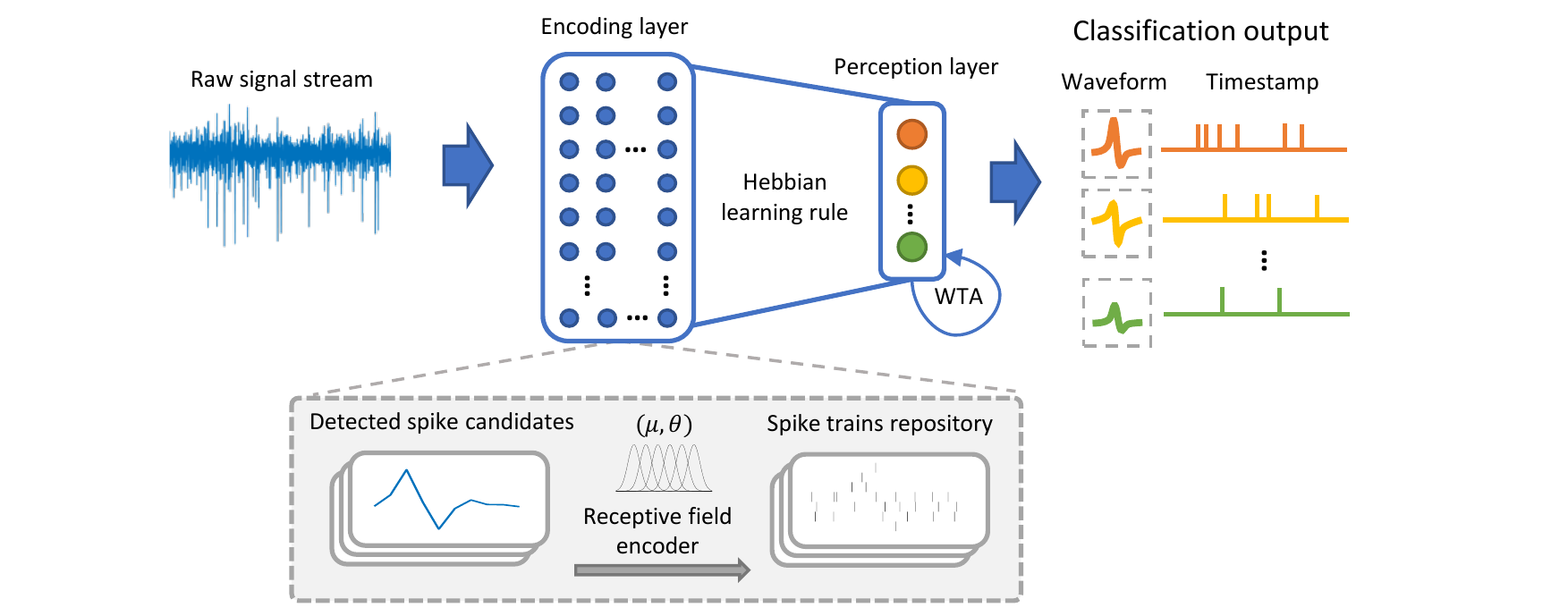}
	\end{center}
	\caption{The overview structure of NeuSort. 
	The raw input signals are firstly preprocessed, then the spike candidates are detected for spike sorting. 
	The encoding layer encodes the waveforms of spikes with a receptive field encoder into event training-based representations, which are fully connected to the perception layer.
	The perception layer distinguishes and assigns the input event trains into different nodes.
	Finally, spike trains are classified and output with exact timestamps.
}
\label{fig_mthd_strct}
\end{figure*}

\section{Method}
\label{section_methods}
Figure~\ref{fig_mthd_strct} depicts the structure of the NeuSort spike sorter for a single channel, which follows the standard pipeline of spike sorting.
Firstly, the raw neural signals undergo preprocessing, and spike candidates are detected.
To classify these spike candidates, the encoding layer employs a receptive field encoder to convert the spike waveforms into spike trains.
The encoding layer is fully connected to the perception layer, where the firing of nodes is based on spiking events, and the weights are updated using the Hebbian learning rule.
Finally, the sorted spikes are output along with their timestamps.
It should be noted that the module presented in Figure~\ref{fig_mthd_strct} represents a single neural signal channel. For multiple channels, each channel has its independent module.

Since the terms "spike" and "neuron" are commonly used in both neural signals and spiking neural networks, in the following content, we will refer to "spikes" and "neurons" in the spiking neural network as "events" and "nodes," respectively.

\subsection{Signal preprocessing}
The conversion of the original raw data into spike candidates involves three steps: filtering, spike detection, and realignment.
The spike candidates are defined as uniformly-sized data segments that contain potential neuronal waveforms or noise. 
These spike candidates are then used as input for the subsequent network.

\subsubsection*{Filtering}
The continuous multi-channel data is initially subjected to band-pass filtering using a 3rd-order Butterworth filter.
This filtering step effectively removes the low-frequency components of the signals, commonly referred to as local field potentials (LFPs), within the frequency range of 0.1-300 Hz.

\subsubsection*{Spike detection}
After spike detection, the raw input signal is segmented into individual spike candidates, which represent significant pulses or changes in the signal.
This segmentation process facilitates the analysis and processing of the signal data.
To capture changes in the energy of the signal, a nonlinear energy operator (NEO) is applied.
In discrete time, the NEO $\psi$ on the filtered signal $S$ is defined as follows:
\begin{equation}
\label{eqt_neo2}
	\psi[S(t)] = S^2(t) - S(t-1)\cdot S(t+1),
\end{equation}
where $S(t)$ is the sample point of the waveform at time $t$.
NEO is designed to be large when the input signal exhibits high power and frequency simultaneously, indicating the presence of a spike \cite{mukhopadhyay1998new}.
This is achieved by comparing the square of the signal amplitude $S^2(t)$ to the product of neighboring samples $S(t+1) \cdot S(t-1)$.
The NEO implementation is relatively simple, utilizing a convolution kernel or a running window, and imposes minimal computational overhead, making it suitable for online applications.
Following spike detection, each spike candidate $C$ comprises $N$ data points.

\subsubsection*{Spike re-alignment}
The purpose of spike alignment is to align each spike candidate $C$ with its maximum amplitude point (which may correspond to the minimum value of the waveform in some cases).
To address potential spike misalignments caused by low sampling rates \cite{Quiroga2004}, the location of the spike candidate maximum is refined using cubic spline-interpolated waveforms iteratively for 5 times.
After the alignment process, the waveforms are downsampled to match the length of the original data points.
It is important to note that accurately measuring the peak of a waveform can be challenging due to its short duration.
As a result, the peak may fall between the sampled time points.

\subsection{NeuSort: the neuromorphic model}
A two-layer neuromorphic network is designed to transform and classify the raw input candidates into distinct clusters, as depicted in Figure~\ref{fig_mthd_strct}.
The initial layer, known as the \textit{encoding layer}, consists of a collection of sensor nodes. 
The subsequent layer is the \textit{perception layer}, comprising a set of Integrate-and-Fire (IF) output nodes.

\subsubsection{Encoding layer}
The encoding layer transforms the spike candidate into a representation of artificial event trains, capturing the waveform shape within the sliding window at each time step.

The Gaussian receptive field, as described in \cite{thorpe1998rank}, enables the mapping of real-valued vectors into a sequence of events.
The receptive field comprises a set of sensor nodes with overlapping sensitivity scales.
The intersection points with each Gaussian node are calculated and converted into events corresponding to a continuous value.
A single continuous value may activate multiple nodes in the field. 
Unlike in \cite{Schliebs2013}, the indices of the activated nodes in the field are retained, while the firing time is disregarded.
\begin{figure}[htbp]
\begin{center}
  \includegraphics[width=0.48\textwidth]{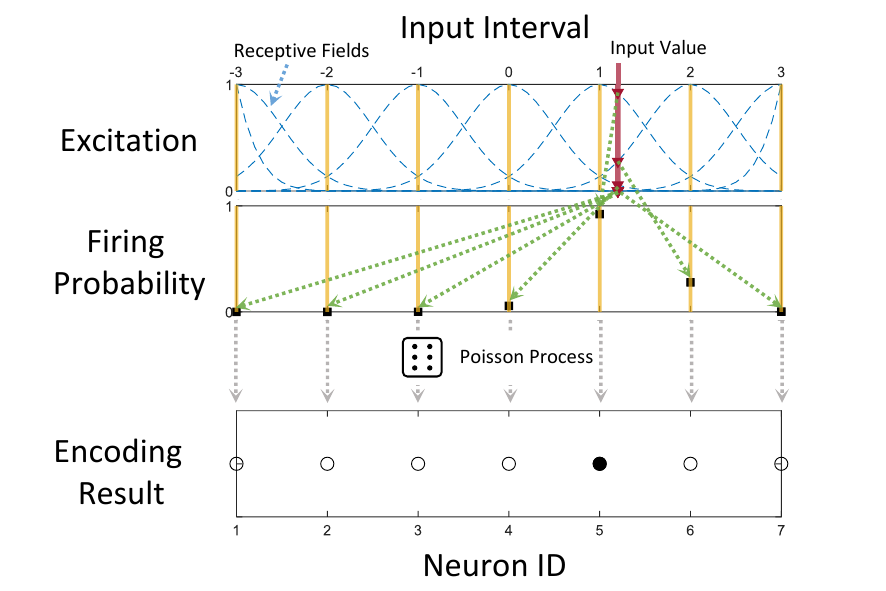}
\end{center}
\caption{
	The encoding process of Gaussian receptive fields.
	For an input value of 1.2 (represented by the red line in the top figure), the intersection points with each Gaussian node are calculated (shown as triangles). 
	These intersection points are then transformed into event firing probabilities (represented by rectangles in the middle figure).
	The probabilities obtained are further converted into the final event output (depicted as circles) using a Poisson process (illustrated in the bottom figure).
}
\label{fig_encode} 
\end{figure}

A spike candidate, denoted as $C_t = [c_{t_1}, c_{t_2}, ..., c_{t_n}]$, represents a sequence of time points from $t_1$ to $t_n$.
Each input variable is independently encoded by a group of sensor nodes with $M$ one-dimensional receptive fields. 
For each $c_{t_n}$, an interval $[I^{c_{t_n}}{min}, I^{c{t_n}}{max}]$ is defined.
The event $J$ of sensor node $\epsilon$ at time $t$ and position $(m, n)$ is calculated as:
\begin{equation}
	J_{{\epsilon}_{(t,m,n)}} = \mathbb{P}_{\mu,\theta}(c_{t_n}),
\end{equation}
where $\mathbb{P}$ represents the Poisson process of the Gaussian receptive field.
The center $\mu_i$ of node $i$ in the Gaussian receptive field is computed as:
\begin{equation}
	\mu_i = I^{C_t}_{min} + \frac{2i-3}{2} \cdot \frac{I^{C_t}_{max} - I^{C_t}_{min}}{M-2}
\end{equation}
and the width $\theta$ is given by:
\begin{equation}
	\theta = \frac{1}{\beta} \cdot \frac{I^{C_t}_{max} - I^{C_t}_{min}}{M-2},
\end{equation}
where $\beta \in [1,2]$ controls the width of each Gaussian receptive field.
Figure~\ref{fig_encode} demonstrates an example of encoding a single variable using the Gaussian receptive field.

During a short time window, the waveforms of spikes from a particular neuron can be considered stationary.
However, due to the stochastic nature of the Poisson process, the same input sequence may result in slightly different outputs, denoted as $I_{{\epsilon}_{(t,m,n)}}$.
By employing a well-designed receptive field, the resulting outputs tend to be similar and can be effectively grouped into the same class.

Receptive field coding plays a crucial role in distinguishing waveform deformations and preventing the confusion of different cells.
Ideally, the distance between the centers of neighboring nodes is slightly larger than the deformation amplitude of adjacent input spikes.
This allows the distortion to be captured by both neighboring nodes in the receptive field and reflected as different weights connected to the next layer.

\subsubsection{Perception layer.}
The role of this layer is to assign artificial inputs to different nodes corresponding to the waveforms of biological neurons.

The output node $\zeta$ represents an Integrate-and-Fire (IF) neuron, where the membrane potential $z$ at time $t$ is governed by the following equation:
\begin{equation}
\label{equ_vol_layer}
	z_{(t,\zeta)} = \sum_n^N \sum_m^M J_{{\epsilon}_{(t,m,n)}} \cdot \omega_{(t,m,n,\zeta)}
\end{equation}
Here, $J_{{\epsilon}{(t,m,n)}}$ denotes the input event series from the encoding layer, and $\omega{(t,m,n,\zeta)}$ represents the fully connected weight from the encoding layer to node $\zeta$ in the perception layer.
The classification output sequence is determined by a predefined threshold $th_d$:
\begin{equation}
	\label{equ_output}
		J_{\zeta_{(t)}}=\Gamma(z_{(t,\zeta)})
\end{equation}
Here, $\Gamma$ represents the comparison between the voltage $z$ and the threshold $th_d$.
When the voltage $z$ exceeds the threshold $th_d$, the output node fires an output spike.

\subsection{NeuSort: online learning}
The learning process of the NeuSort network can be described as follows.
\begin{itemize}
	\item Initial state:
	The output nodes study the input waveforms and differentiate them.
	Since there is no corresponding output node for the event train input, a randomly selected output node is updated and does not generate event output.
	
	\item Deformation state:
	The output nodes track the deformed neurons and update themselves.
	For event train input, the corresponding output node is selected using the Winner-Takes-All (WTA) mechanism and fires.
	The connection weights between the two layers are updated using the Hebbian learning rule.
\end{itemize}	

The Hebbian learning rule is applied for updating the neuromorphic network, based on the principle that neurons fired together should wire together \cite{hebb1955drives}.
This rule allows the network to learn stable patterns of event activity over time while ignoring noise that lacks stable patterns.
When a waveform, encoded as artificial event trains, is input to the encoding layer, a specific group of nodes is activated in response.
The connection weights $\hat{\omega}$ between the two layers are updated using a constant value $\tau_{{h}{+}}$ or $\tau{{h}{-}}$ according to the following rule:
\begin{equation}
\label{equ_stdp}
\dot\omega_{(m,n,\zeta)}=
\cases{
	\min(\omega_{(m,n,\zeta)} + \tau_{{h}_{+}}, \omega_{max}) & if in events,\\
	\max(\omega_{(m,n,\zeta)} - \tau_{{h}_{-}}, \omega_{min}) & else.\\}
\end{equation}

As the waveform is repeatedly presented, a certain group of nodes are activated in accordance with the Hebbian rule to "remember" the waveform pattern, thus enabling the network to learn to detect the waveform.
In this way, the NeuSort spike sorter can automatically learn to sort waveforms, and all the signals pass through the network only once.

To ensure that each biological neuron is uniquely represented by a single node in the perception layer, a winner-take-all (WTA) mechanism is employed during the learning process. 
In our implementation, the node selection in the perception layer follows a WTA mechanism:
\begin{equation}
	\dot\zeta = \max_{\zeta} z_{(t,\zeta)}
\end{equation}
Here, $\dot{\zeta}$ denotes the selected node that undergoes updates. 
If no node is selected, the incoming spike is randomly assigned to a node in the perception layer. 
Specifically, when one node is activated, the other nodes are inhibited and their updates are suppressed. 
Only the active node strengthens its connections to learn the input pattern. 
Consequently, when a similar waveform segment occurs, the active node is more likely to fire, thus reinforcing its specificity in responding to that particular waveform pattern.

The WTA mechanism enables adaptive tracking of neurons against waveform deformation. 
As the waveform of a neuron undergoes slight changes, the corresponding node in the perception layer remains the same, allowing the network to adapt and adjust to the altered waveform patterns. 
When the waveform undergoes abrupt changes, indicating the presence of a new neuron, a new node is assigned in the perception layer to automatically detect the arrival of new neurons.

In our implementation, the output trains directly correspond to the spike firing times in the input stream, while other methods rely on output queues to determine the ownership of input spikes through voting.

The NeuSort spike sorter operates in a fully automatic manner without requiring supervision. 
As unsorted spike waves are continuously fed into the network, NeuSort progressively learns to converge towards an effective spike sorter by "remembering" the repeatedly occurring patterns (neuronal waveforms) with specific groups of nodes, following the synaptic weight update rule of the neuromorphic model. 
The synaptic updates track and adapt to changes in spike waveforms.

\subsection{Criteria}
Though various methods have been proposed, the agreement among them for the same high-density dataset is often low, primarily due to false positives \cite{Buccino2020}. 
When comparing a sorting output to a ground truth, each spike can be categorized as:
\begin{itemize}
	\item True positive (\textit{tp}): Spikes found in both the ground truth and the output.
	\item False Positive (\textit{fp}): Spikes found in the ground truth, but not in the output.
	\item False negative (\textit{fn}): Spikes found in the output, but not in the ground truth.
\end{itemize}

In addition, the following performance measures are calculated:
\begin{itemize}
	\item Accuracy: ${\#tp}/({\#tp+\#fn+\#fp})$
	\item Recall: ${\#tp}/({\#tp+\#fn})$
	\item Precision: ${\#tp}/({\#tp+\#fp})$
\end{itemize}

Apart from these conventional assessments, a global F-score $F$ is computed across all ground truths and outputs using the following formula:
\begin{equation}
	F = \frac{2 * H}{T + O},
\label{equ_fscore_ori}
\end{equation}
Here, $T$ represents the total number of ground truths, $O$ represents the total number of outputs, and $H$ is the number of outputs that coincide with the ground truths.

Using the aforementioned parameters, Equation~\ref{equ_fscore_ori} can be rewritten as: $F = ({2*\#tp})/({2*\#tp+\#fn+\#fp})$.

\section{Experimental results}
\label{section_experimental_results}
In the results section, we validated the performance of our proposed method using two real-world datasets. 
We compared the results with other algorithms, such as PCA-KM\cite{Adamos2008}, Wave\_clus\cite{Chaure2018}, Osort\cite{Rutishauser2006}, HerdingSpikes2\cite{HILGEN20172521}, IronClust\cite{jun2017fully}, Kilosort2\cite{Pachitariu2016}, Mountainsort4\cite{Chung2017}, and Tridesclous\cite{Garcia2015}. 
Additionally, we evaluated the computational costs associated with each algorithm and provided insights into the learning process of the network and the parameter setting procedure. 
We also estimated the computational overhead on a neuromorphic chip. 
Furthermore, we conducted experiments on simulated datasets and compared the results with those obtained from other comparative methods. 
For detailed simulation results, please refer to the appendix.

All experiments were conducted using MATLAB R2020a (MathWorks, Natick, MA) on a personal computer equipped with an Intel Core i5-10400 2.90 GHz CPU and 48GB RAM.

\subsection{Datasets and comparative methods}
\subsubsection{Description of the datasets}
Two real neural benchmark datasets were used in the experiments.
The first dataset included ground truth recordings containing neurons with waveform deformations.
The second dataset consisted of ten distinct segments of long-duration multi-channel recordings.

\begin{itemize}
	\item \textbf{dataset\_real1.}
		The first real dataset is available from \cite{Henze2000,henze2009}, which provides only a single known ground truth neuron. The recordings are recorded from the CA1 hippocampal region of anesthetized rats, which contain the same neuron recorded both intracellularly and extracellularly.
		One example (dataset d553101) of the scaling of the ground truth waveform over time is found in the dataset, which can be caused by the distance shift between the extracellular probe and the neuron during the acquisition process. In Figure~\ref{fig_rslt_real_crcn}A, the ground truths are stained, ranging from cyan to magenta on the timeline. The first 100 spikes of ground truths (the left plot of Figure~\ref{fig_rslt_real_crcn}B) and the last 100 ones (the middle plot in Figure~\ref{fig_rslt_real_crcn}B) are identified easily as two neuron waveforms. From the perspective of waveforms, it is difficult to cluster the spikes of ground truths at various time points as the exact one.  
	\item \textbf{dataset\_real2.}
		Another real dataset contains neural signal data recorded with the Utah array (with 96 channels) from the motor cortex of a rhesus when performing a four-direction center-out task \cite{wang2014neural}. We use ten data segments recorded from different days (from 'MC01' to 'MC10'). Skilled sorters perform manual spike sorting to provide the label as the ground truth. 
		All spike candidates, along with their corresponding timestamps, were pre-detected in the original data and stored in .mat files.
\end{itemize}

\subsubsection{Description of the comparative methods}
We compare our method with three representative approaches through visual analysis: PCA-KM\cite{Adamos2008}, Wave\_clus\cite{Chaure2018}, and Osort \cite{Rutishauser2006}. 
Furthermore, we evaluate our approach against five state-of-the-art (SOTA) sorters: HerdingSpikes2\cite{HILGEN20172521}, IronClust\cite{jun2017fully}, Kilosort2\cite{Pachitariu2016}, Mountainsort4\cite{Chung2017}, and Tridesclous\cite{Garcia2015}, using statistical measures.
The parameters in all the compared automatic methods are set to the corresponding default values.

\begin{itemize}
	\item \textbf{PCA-KM.}
	PCA-KM is a classical spike sorting approach that utilizes principal component analysis (PCA) and K-means for feature extraction and clustering, respectively.
	To establish the initial principal components, a portion of the data is used as the training set.
	Due to the limited number of distinguishable neurons in a channel (not exceeding three)\cite{Rey2015}, we set $K=3$ in the following experiments.
	\item \textbf{Wave\_clus.}
	Wave\_clus is a spike sorting algorithm that employs wavelet decomposition to extract features from spike waveforms and superparamagnetic clustering (SPC) to cluster the spikes in the feature space.
	For comparison purposes, we substitute PCA as the feature extraction method and fix the temperature parameter $T$ in SPC for online processing.
	\item \textbf{Osort.}
	Osort is an algorithm that utilizes valley-seeking clustering and offers a favorable accuracy-complexity tradeoff.
	The source code of Osort is provided for testing, and we use the code with the parameters recommended in their article.
	Some of the detection parameters are adjusted to achieve better performance.
	\item \textbf{HerdingSpikes2.}
	HerdingSpikes starts with fast spike detection on individual channels, optimizing the performance further with interpolation-based spike detection. 
	Spatial spike localization is then employed to enable higher-resolution analysis, followed by spike sorting based on the combined information of spike locations and waveform features.
	\item \textbf{IronClust.}
	IronClust specially addresses probe drift during spike sorting by tracking anatomical similarity between time chunks, linking them based on proximity, and constraining the spike neighborhood within linked chunks.
	\item \textbf{Kilosort2.}
	KiloSort is a template matching-based package for high-density electrode array online applications. It reduces manual curation through spatial masking, template iteration, and post-hoc merging.
	\item \textbf{Mountainsort4.}
	Mountainsort operates on the principle of performing spike sorting separately on neighborhoods, defined as feature spaces around a central electrode.
	The approach does not assume specific cluster distributions and aims to consolidate clusters to reduce redundancies.
	\item \textbf{Tridesclous.}
	Tridesclous is a template matching based mathod. It consists of constructing a catalog through automated preprocessing, waveform analysis, feature extraction, and clustering. The catalog is then manually checked and corrected. After construction, it can be performed offline or in real-time.
\end{itemize}

\subsection{Performance on the real datasets}
We assess the performance of NeuSort using two real datasets: one consisting of single-channel intracellular recordings and another comprising multi-channel neural data collected on different days.

The spike sorters were evaluated in a pseudo-online fashion, where the spike waveforms were sequentially presented to the spike sorters. 
In the case of PCA-KM, a training set comprising stable spike waveforms was utilized for parameter calibration. However, Wave\_clus, Osort, and NeuSort did not require any training procedures.

\begin{figure*}[htbp]
	\begin{center}
		\includegraphics[width=\textwidth]{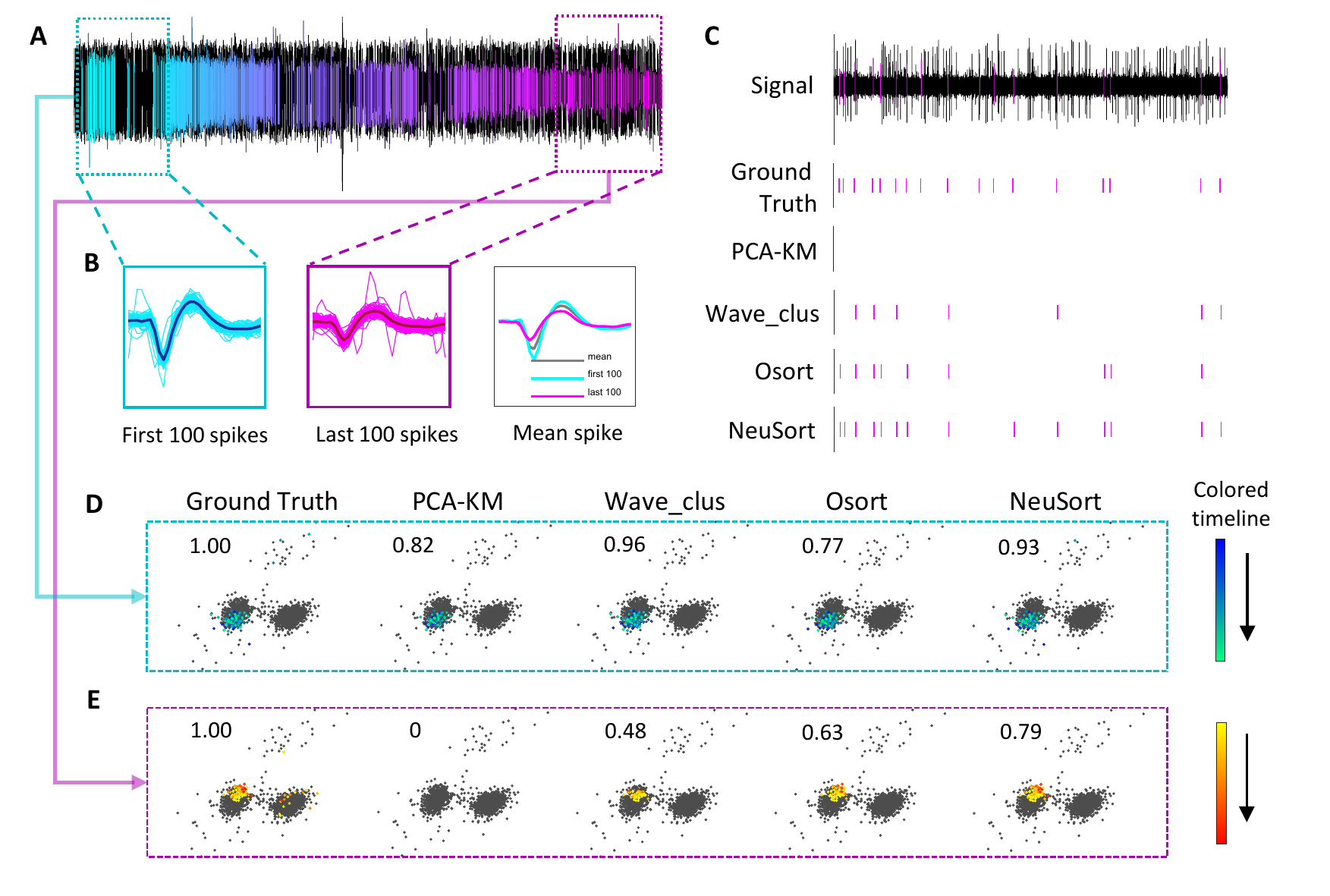}
	\end{center}
	\caption{Performance with dataset\_real1. 
		(A) The ground truths obtained from intracellular recordings, color-coded from cyan to magenta.
		(B) The first 100 spikes and the last 100 spikes are displayed in cyan and magenta, respectively. The average waveforms of all spikes, as well as the first and last 100 spikes, are presented.
		(C) Spike sorting results of different methods for one segment.
		(D) \& (E) Spike sorting results for the first and last 100 spikes. The temporal evolution of the spiking events is indicated by different colors, as shown in the colorbar on the right. The F-scores are indicated at the top left.
	}
\label{fig_rslt_real_crcn} 
\end{figure*}

\subsubsection{Performance with single probe intracellular recordings}
We evaluated the performance of NeuSort under the condition of waveform deformation using a real dataset. 
The dataset consists of intracellular cell recordings, which can be considered as the ground truth. 
The ground truths are color-coded from cyan to magenta, as shown in Figure~\ref{fig_rslt_real_crcn}A. 
Although all ground truths should belong to a single cluster based on intracellular records, it can be observed that the ground truths exhibit separate properties during different periods, as indicated by the cyan and magenta spikes in Figure~\ref{fig_rslt_real_crcn}B.

The spike sorting performance is evaluated using the F-score. 
The F-score of the ground truth is defined as 1. 
In the initial spikes, as shown in Figure~\ref{fig_rslt_real_crcn}D, all four spike sorters achieve high spike sorting performance. 
Specifically, the F-scores are 0.82, 0.84, 0.83, and 0.90 for PCA-KM, Wave\_clus, Osort, and NeuSort, respectively. 
However, as the spike waveform gradually changes over time, static spike sorters like PCA-KM lose track of the neuron. 
In the last part of the recording, the F-score of PCA-KM becomes 0. 
On the other hand, NeuSort maintains a high F-score of 0.79 under the waveform deformation condition, which is 0.31 and 0.16 higher than Wave\_clus and Osort, respectively.
Figure~\ref{fig_rslt_real_crcn}C illustrates the spike detection results of different spike sorters for a segment at the end of the recording. 
The results demonstrate that NeuSort can accurately capture neuronal firing behaviors and outperform other methods.

We conducted a comparative analysis using the SpikeForest framework \cite{magland2020spikeforest} and compared the performance of our method against other state-of-the-art algorithms. 
Table~\ref{tab_method_cmp_on_real1} displays the comprehensive classification outcomes, where NeuSort demonstrates the highest number of \#tp while simultaneously minimizing the occurrence of \#fp. 
As a result, our approach achieves the highest accuracy of 0.82 and precision of 0.91 on this specific dataset.

\begin{table*}[htbp]
\begin{center}
\caption{Performance comparison with dataset\_real1.}
\label{tab_method_cmp_on_real1}
\begin{threeparttable}
\begin{tabular}{l | c ccc | ccc}
	\toprule
	& \makecell{\textbf{\# spike} \\ \textbf{detected}} & 707 & & & & \\
	\midrule
	\textbf{Method} & \makecell{\textbf{\# spike} \\ \textbf{assigned}} & \textbf{\# tp} & \textbf{\# fn} & \textbf{\# fp} & \textbf{Accuracy} & \textbf{Precision} & \textbf{Recall} \\
	\midrule
	PCA-KM & 551 & 484 & 156 & 67 & 0.68 & 0.76 & 0.88 \\
	Wave\_clus & 630 & 513 & 77 & 117 & 0.73 & 0.87 & 0.81 \\
	Osort & 651 & 502 & 56 & 149 & 0.71 & 0.90 & 0.77 \\
	HerdingSpikes2 & 686 & 499 & 83 & 104 & 0.73 & 0.86 & 0.83 \\
	IronClust & 687 & 520 & 122 & 45 & 0.76 & 0.81 & 0.92 \\
	Kilosort2 & 703 & 531 & 79 & 93 & 0.76 & 0.87 & 0.85 \\
	Mountainsort4 & 663 & 511 & 115 & 37 & 0.77 & 0.82 & 0.93 \\
	Tridesclous & 690 & 509 & 138 & 43 & 0.74 & 0.79 & 0.92 \\
	\textbf{NeuSort(ours)} & 649 & 578 & 58 & 71 & 0.82 & 0.91 & 0.89 \\
	\bottomrule
\end{tabular}
\end{threeparttable}
\end{center}
\end{table*}


\begin{figure*}[htbp]
  \begin{center}
    \includegraphics[width=\textwidth]{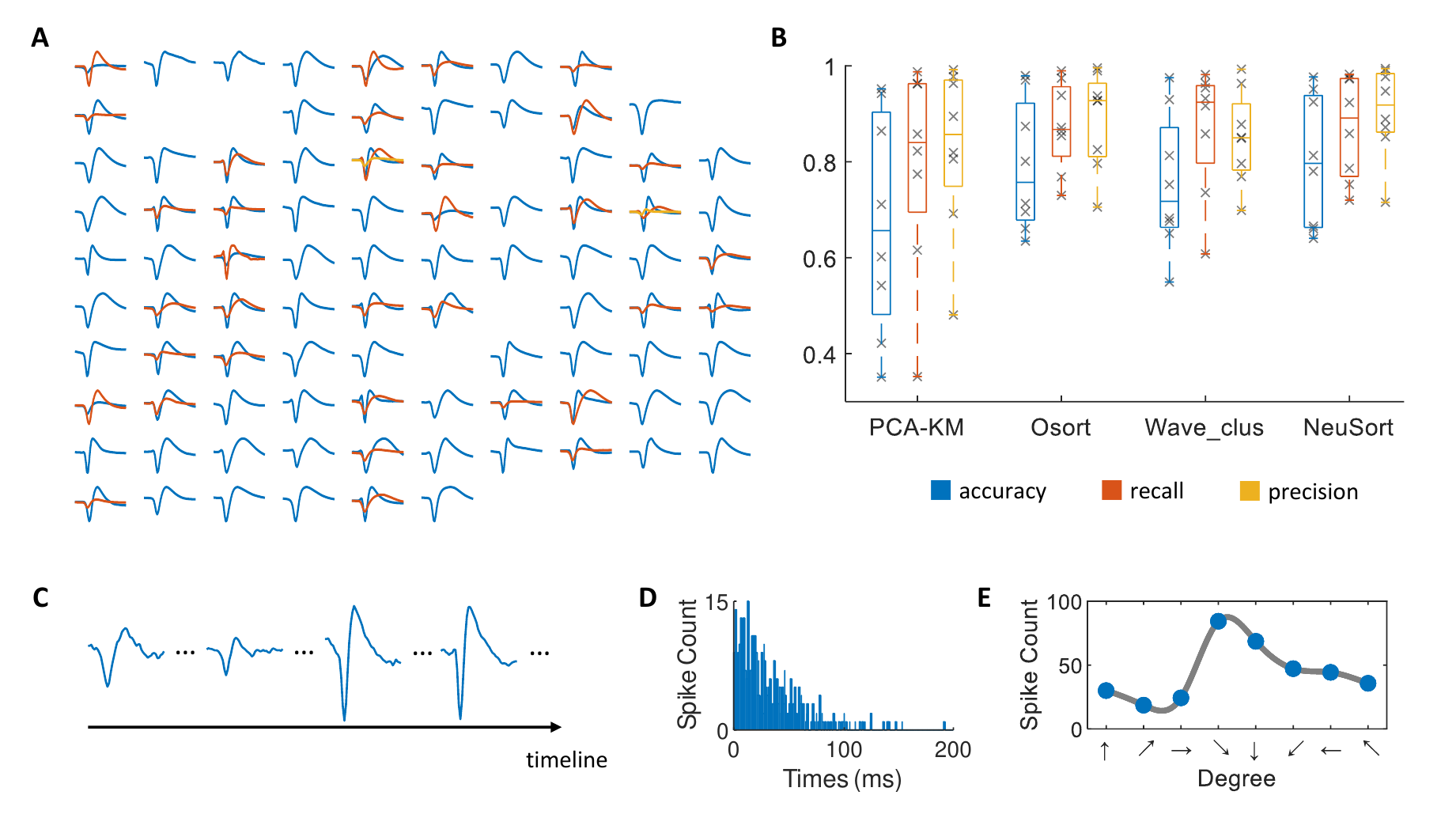}
  \end{center}
  \caption{
    (A) Average waveforms detected from neural signals recorded by a 96-channel Utah array.
    (B) Performance comparison using boxplot on dataset\_real2.
    (C) Temporal waveform deformations of a putative neuron.
    (D) Inter-Spike interval (ISI) analysis of the same putative neuron.
    (E) Tuning curve analysis of the same putative neuron.
  }
\label{fig_stat_method_cmp_on_realdata}
\end{figure*}

\subsubsection{Results on multi-day collections}
We evaluate the performance of NeuSort on a larger real dataset obtained from Utah arrays (96 channels) over multiple days (MC01 to MC10). 
Ground truth labels are established through manual spike sorting conducted by experienced experts.
Figure~\ref{fig_stat_method_cmp_on_realdata}A displays the average waveforms detected by NeuSort for each channel in dataset MC01, with channel placement corresponding to the Utah array configuration.
Most channels exhibit one or two putative neurons, while some channels have no neurons or contain three neurons.

A quantitative comparison of spike sorting performance is shown in Figure~\ref{fig_stat_method_cmp_on_realdata}B. 
In this multi-channel dataset, NeuSort achieves the highest mean accuracy of 0.79 and mean precision of 0.89, outperforming the other compared sorters and demonstrating its capability to process large amounts of data effectively.
Specifically, Osort, Wave\_clus, and NeuSort exhibit average accuracies of 0.77, 0.75, and 0.79, respectively, and precisions of 0.87, 0.84, and 0.89, respectively. 
The four sorters show similar mean recalls of 0.85, 0.87, 0.87, and 0.86 for PCA-KM, Wave\_clus, Osort, and NeuSort, respectively. 

Notably, waveform deformations are observed in some high-performing datasets, as depicted in Figure~\ref{fig_stat_method_cmp_on_realdata}C. 
The waveform of the putative neuron exhibits significant deformations. 
Statistical analysis reveals an average interspike interval (ISI) of 33.28 ms for this neuron (Figure~\ref{fig_stat_method_cmp_on_realdata}D), indicating its sensitivity to motion in the downward right direction (Figure~\ref{fig_stat_method_cmp_on_realdata}E). 
The evaluation metrics pertaining to the identified neurons are available in Section~\ref{dis_deform}

Furthermore, we conducted a comparative analysis with SOTA spike sorting methods.
The F-score and accuracy metrics were calculated by averaging the results across all data segments. 
Our method achieved an F-score of 0.86$\pm$0.10 and an accuracy of 0.79$\pm$0.16, placing it at an intermediate level of performance compared to the other algorithms examined. 
Table~\ref{tab_cmp_si} provides the complete F-score and accuracy values for each algorithm. 

\begin{table}[htbp]
  \begin{center}
    \caption{Performance comparison with dataset\_real2.}
    \label{tab_cmp_si}
    \begin{threeparttable}
    \begin{tabular}{l | c c}
      \toprule
      \textbf{Algorithm} & \textbf{F-score} & \textbf{Accuracy} \\
      \midrule
	  PCA-KM & 0.82$\pm$0.25 & 0.72$\pm$0.27 \\
	  Wave\_clus & 0.86$\pm$0.11 & 0.77$\pm$0.16 \\
	  Osort & 0.85$\pm$0.10 & 0.75$\pm$0.15 \\
      HerdingSpikes2 & 0.64$\pm$0.22 & 0.48$\pm$0.28 \\
      IronClust & 0.88$\pm$0.10 & 0.79$\pm$0.16 \\
      Kilosort2 & 0.89$\pm$0.10 & 0.81$\pm$0.17 \\
      Mountainsort4 & 0.76$\pm$0.21 & 0.59$\pm$0.28 \\
      Tridesclous & 0.75$\pm$0.16 & 0.60$\pm$0.19 \\
      \textbf{NeuSort (Ours)} & 0.86$\pm$0.10 & 0.79$\pm$0.16 \\
      \bottomrule
    \end{tabular}
    \end{threeparttable}
  \end{center}
\end{table}

\subsection{Analysis of the learning process}
We validate the performance of NeuSort by tracing the synapse weights of the model during one learning process, both in the initialization stage and the online adaptation of waveform changes. 
For detailed information about the simulated datasets used in the analysis of the learning process, please refer to the appendix.

\begin{figure*}[htbp]
	\centering
	\includegraphics[width=\textwidth]{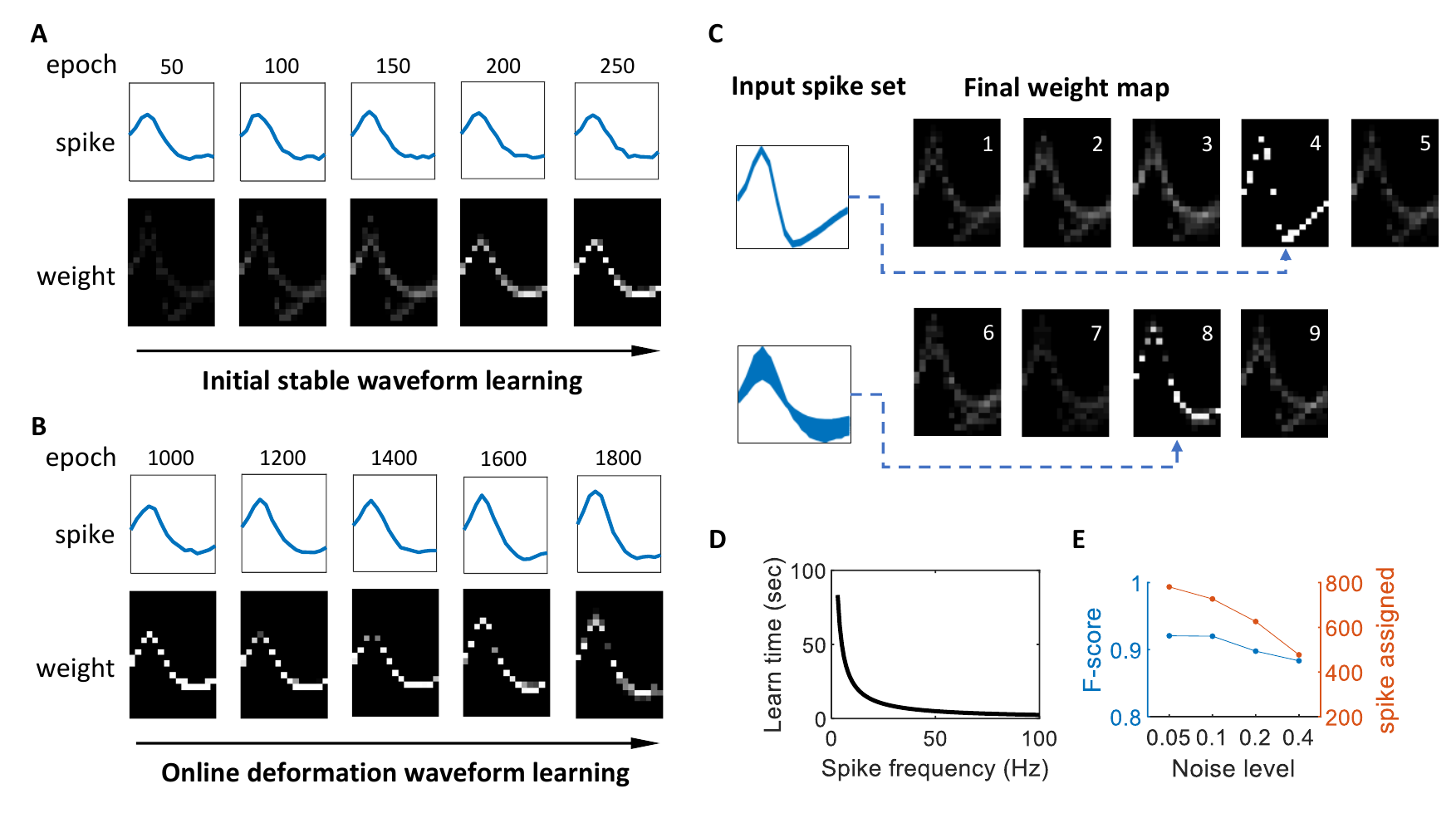}
	\caption{The learning process of NeuSort. 
		(A) The learning process with steady waveforms.
		(B) The automatic tracing with deformation waveforms.
		(C) The learned weight maps of the network.
		(D) Time of convergence against the firing rate of neurons.
		(E) Quantification of F-score and spike assignment across varied noise levels.
	}
\label{fig_STDP_learn} 
\end{figure*}

\subsubsection{The initial learning process}
Firstly, we examine the plug-and-play learning process of NeuSort. 
As depicted in Figure~\ref{fig_STDP_learn}A, the first row illustrates the spike waveform at each epoch step (referred to as time $t$), and the second row illustrates the synapse weight map $\omega$. 
The weights are initialized to zero. 
As the model receives the same waveform repeatedly, the pattern gradually emerges in the synapse. 
Around the 200$^{th}$ epoch, the model has learned the waveform.

Since the weight map updating is triggered by the appearance of spikes, neurons with higher firing rates require less time for spike sorter training. 
Figure~\ref{fig_STDP_learn}D demonstrates the time required to obtain a stable sorter under different firing rates. 
NeuSort mainly converges in less than 100 seconds. 
Due to its fully automatic and unsupervised nature, it can serve as a plug-and-play pipeline for online spike sorting.

\subsubsection{Online adaptation with deformation waveforms}
In Figure~\ref{fig_STDP_learn}B, we investigate the online adaptation ability of NeuSort with waveform deformations. 
As shown in the first row, the spike waveform starts to deform from the 1000$^{th}$ epoch. 
The corresponding synapse weight map in the second row effectively tracks the changes in the waveform over time.

With gradual and continuous waveform deformations, the same nodes in the perception layer are activated due to the overlap between adjacent receptive field nodes. 
Consequently, the nodes learn to adapt to waveform changes, and the weights corresponding to the deformed spikes are updated (weight maps from the 1200$^{th}$ epoch to the 1800$^{th}$ epoch). 
The results demonstrate that NeuSort exhibits adaptability to online waveform deformations, thereby enhancing the robustness of long-term spike sorting tasks.

In Figure~\ref{fig_STDP_learn}C, we present the final synapse weight map after the initial learning and online adaptation phases. 
Among the nine weights, only weights 4 and 8 exhibit evident and meaningful spike patterns, indicating the presence of two distinct neurons in the channel. 
Through the winner-takes-all (WTA) strategy, the network achieves effective regulation and tends to utilize a single synapse weight map to represent a waveform pattern. 
The "blank" synapse weight maps are preserved to capture incoming waveforms.

To provide a more robust assessment under challenging and realistic conditions, we curate a hybrid dataset encompassing varying degrees of background noise. 
This hybrid dataset is formulated by incorporating actual distorted spikes from \textit{dataset\_real1} and augmenting them with background noise at different Signal-to-Noise Ratios (SNRs), as elucidated in the appendix.
The noise level is ascertained based on the standard deviation of the noise amplitude.
In Figure~\ref{fig_STDP_learn}E, we present a graphical representation of NeuSort's F-score and spike assignment performance across varying levels of noise. 
As the noise levels escalate, the count of classifiable spikes diminishes, ranging from 781 (noise level=0.05) to 447 (noise level=0.4). 
This decline in the count of classifiable spikes can be primarily attributed to the heightened impact of noise during the spike detection phase.
With increasing noise levels, a greater proportion of candidate spikes becomes intertwined with noise, thus leading to a decrease in the number of samples subsequently entering the clustering process.
This phenomenon underscores an inherent limitation prevalent in existing spike detection methodologies.
Despite this trend, the F-score exhibits only a modest reduction, descending from 0.92 (noise level=0.05) to 0.88 (noise level=0.4). 
This observation underscores the continued efficacy of our methodology in discerning candidate spikes amidst high levels of noise.
To further broaden the comparative analysis of NeuSort with alternative methodologies on the hybrid dataset, we encourage reference to the appendix for additional insights.

\subsection{Parameter settings}
The parameter settings of NeuSort are summarized in Table~\ref{tab_encoding_para}.
During the spike sorting process, all spike candidates undergo filtering using a passband of [300, 3000] Hz. Subsequently, the spike waveforms are cropped to a length of $N=64$.

\begin{table}[thp]
\begin{center}
\caption{Settings of NeuSort hyper-parameters.}
\label{tab_encoding_para}
\begin{tabular}{clc}
	\toprule
	Parameters & Description & Value \\
	\midrule
	$I_{max}$ & The maximum of receptive field & 200 \\
	$I_{min}$ & The minimum of receptive field & -200 \\
	$\beta$ & Field shape factor & 2 \\
	$d_r$ & Mean neighbor receptive field node distance & 13 \\
	$th_{d}$ & Perception node firing threshold & $N\times0.4 $ \\
	$\tau_{h_+}$ & Short-term plasticity time presynaptic constant & 0.2 \\
	$\tau_{h_-}$ & Short-term plasticity time postsynaptic constant & 0.1 \\
	$\omega_{min}$ & Lower weight bound & 0 \\
	$\omega_{max}$ & Upper weight bound & 1 \\
	\bottomrule
\end{tabular}
\end{center}
\end{table}

For the receptive field, the boundary $[I_{min}, I{max}]$ of input time points of spike candidates is set to $[-200, 200]$, and the field shape factor $\beta$ is set to 2.
The number of nodes $M$ is normally specified manually. 
In our approach, we dynamically calculate $M=(I_{max}-I_{min})/{d_r}$ through the bounds of field $[I_{min}, I_{max}]$ and the distance $d_r$ between receptive field nodes.
The value of $d_r$ varies across different datasets depending on the degree of waveform deformation.
The deformation of the waveform should not exceed half of the receptive field width, i.e., $d_r/2$, as it may not be recognized by adjacent nodes.
Specifically, the value of $d_r$ is determined by the signal-to-noise ratio (SNR) of the dataset and is inversely proportional to the SNR.
In real datasets, we uniformly set the value of $d_r$ to 13, resulting in $M=\lceil(I_{max}-I_{min})/d_r\rceil=31$.

For the network structure, the scale of the encoding layer is determined by the node number $M$ of the receptive field, and the length $N$ of input time trains, $N$ varies according to different datasets.
The scale of the perception layer is set to $9\times1$.
The encoding layer is fully connected to the perception layer, and the weights of synapses are initialized to 0.

The firing threshold $th_d$ of the perception layer is determined by the length of the input, denoted as $N$.
Typically, only one node in encoding layer is activated at each time point, ensuring that the accumulated voltage of a node in the perception layer, which is crucial for accurate classification, does not exceed $N$.
Considering that the weights do not need to reach the boundary during the running process, we empirically set the threshold to $N\times0.4$.

\begin{figure}[htbp]
\centering
\includegraphics[width=0.48\textwidth]{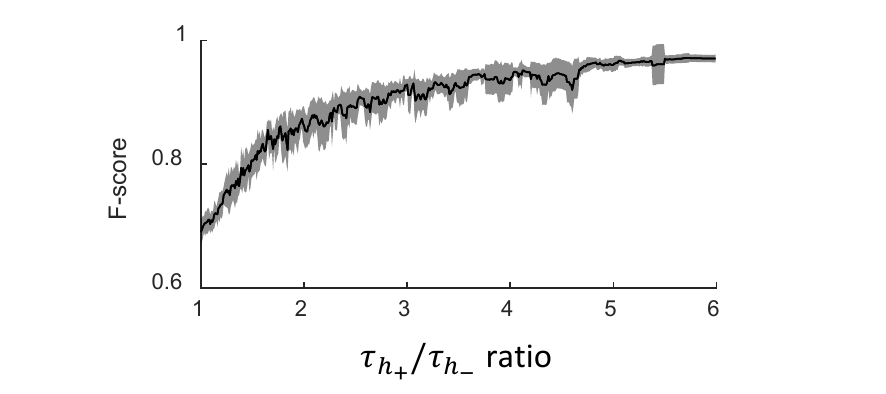}
\caption{Evaluating the influence of parameter ratio of $\tau_{h_+}$ and $\tau_{h_-}$ on dataset\_syn1 ($th_d=0.4\times N, \tau_{h_-}=0.1$).}
\label{fig_mthd_lrn_ratio}
\end{figure}

For weight update, Hebbian learning rules apply on the synapses stemming from the encoding layer to the perception layer for each synaptic spike occurrence.
To ensure the weights stay within a reasonable range, we set the upper and lower bounds of the weights, denoted as $[\omega_{min}, \omega_{max}]$, to $[0,1]$.
The parameter $\tau_{h_+}$ and $\tau_{h_-}$, which represents the learning ratio, plays a crucial role in determining the speed at which the network learns to recognize a fixed waveform. 
A higher learning ratio corresponds to a faster learning rate, enabling the network to quickly identify the waveform. 
However, excessively sharp learning ratios do not significantly improve recognition accuracy and may cause the nodes in the perception layer to solidify too quickly, leading to over-clustering.
To investigate the impact of different parameter ratios, we evaluate them on dataset\_syn1 (detailed in the appendix) and present the effect on recognition accuracy in Figure~\ref{fig_mthd_lrn_ratio}. We observe that the slope of the accuracy curve starts to slow down at around 1.7 on the abscissa.
Taking into account both the recognition speed and accuracy, we ultimately select the parameters $\tau_{h_+}=0.2$ and $\tau_{h_-}=0.1$.

\subsection{Comparison of computational cost}
\label{dis_computation_cost}
The computational time required for spike sorting is a crucial factor, as it should be shorter than the duration of the recording \cite{Buccino2022}. 
This is particularly important for downstream analysis and closed-loop BCI decoding \cite{zaghloul2015implementable,Werner2016,zaghloul2015adaptive}.
In this regard, we analyze the computational cost of different spike sorters using a simulated dataset \textit{dataset\_syn1} with a 10-fold increase. 
The details of the dataset can be found in the appendix.
We leverage 1500 spike samples and record only the operation time.

\begin{figure}[htbp]
\centering
\includegraphics[width=0.48\textwidth]{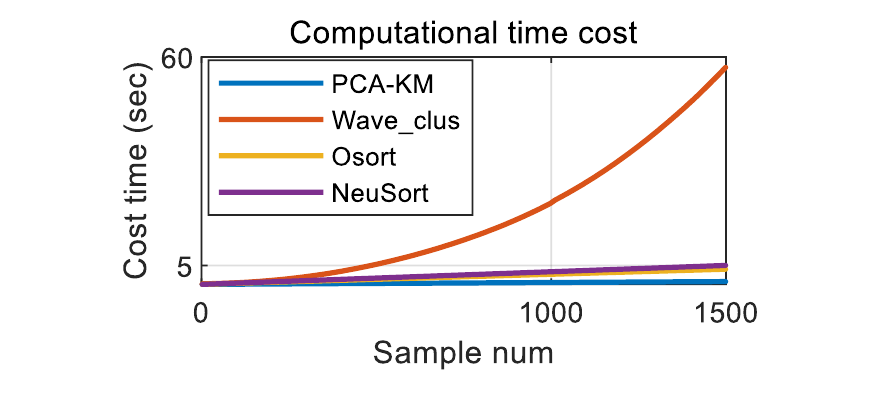}
\caption{Comparison of computational costs.}
\label{fig_rslt_cost_time}
\end{figure}

The results, presented in Figure~\ref{fig_rslt_cost_time}, demonstrate the computational cost of the four sorters. 
Among them, PCA-KM, Osort, and NeuSort exhibit linear computational time as the number of samples increases, with respective times of 0.74, 4.55, and 5.82 seconds. 
PCA-KM requires the shortest time. 
In contrast, the Wave\_clus approach exhibits an exponential increase in computational cost with the growing number of spikes. 
This is due to the sorting process involving all historical data, resulting in a total classification time of 58.35 seconds for all samples.

NeuSort requires slightly more time than Osort because it takes additional time to encode the original input into the spike series. 
The linear growth of computational time exhibited by NeuSort indicates its ability to handle long-term online sorting.

\subsection{Power consumption estimation}
The proposed SNN structure has a significant advantage in terms of high energy efficiency and ultra-low cost on neuromorphic chips. 
To evaluate the power consumption, we refer to previous work on similar neuromorphic hardware.

Based on the findings in \cite{Cao2015}, we assume that the energy emitted by a single spike is denoted as $\alpha$. 
During the online recognition phase, we measure an average of 41 spikes to process a single input waveform in \textit{dataset\_real2}, with a rate of $3.42\times10^3$ inputs per second. 
Thus, the energy consumed for processing each channel is estimated to be $1.4\times10^5 \alpha$.
By multiplying this energy consumption by the number of channels (96), the total power requirement $P$ can be calculated as follows:
\begin{eqnarray}
	P &= 1.4 \times 10^5 \alpha \ Watts/channel \times 96 \ channel\nonumber\\
	  &= 1.34 \times 10^7 \alpha \ Watts
\label{equ_power_estimation}
\end{eqnarray}

Taking into consideration two previously published spike-based neuromorphic circuits, namely Loihi \cite{davies2018loihi} and Neurosynaptic core \cite{arthur2012building}, their respective single spike energy values ($\alpha$) have been documented as 23.6 and 45.0 pJ. 
Substituting these values into Equation~\ref{equ_power_estimation} yields estimated power consumptions of 0.317 and 0.605 mW, respectively. 
This comparison underscores the substantial power consumption advantage of neuromorphic chips over traditional counterparts.

\section{Discussion}
\label{section_discussion}
We propose an online spike sorting method based on a neuromorphic model, which is capable of adapting to online variations in neural signals and the emergence of new neurons.
In our method, the raw signal is processed in a streaming fashion, and each putative neuron output is assigned a corresponding timestamp.
Following each output, the associated nodes and connections are updated to track the changes in neurons.
Therefore, in the presence of neuronal deformations, the nodes within the receptive field exhibit continuous temporal firing changes, and the network dynamically tracks and adjusts the connection weights accordingly.
Additionally, the signal undergoes a single pass through the network, unlike traditional spiking neural networks that require multiple iterations, thereby effectively reducing computational energy consumption.

\begin{figure*}[htbp]
\centering
\includegraphics[width=\textwidth]{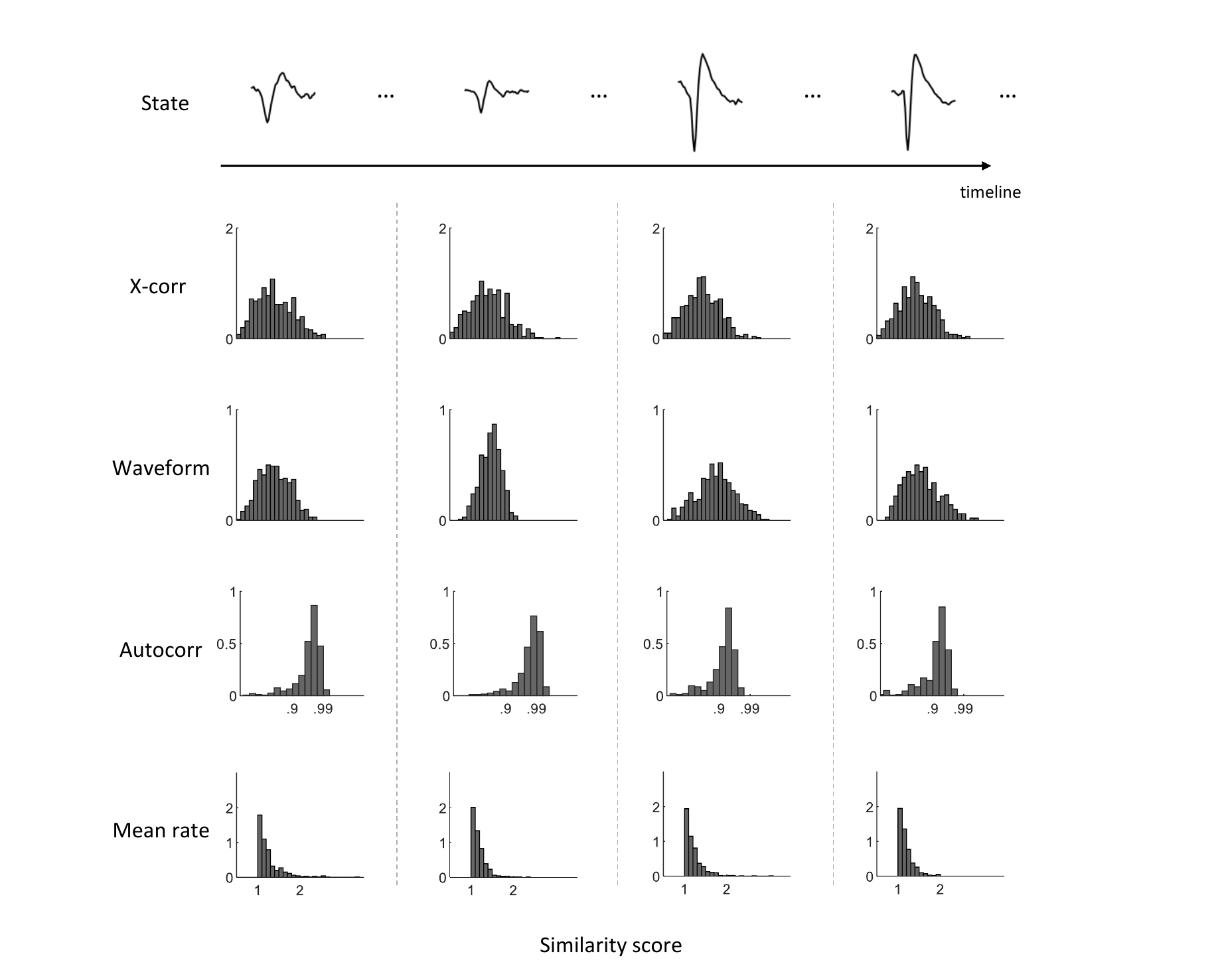}
\caption{
	Distributions are shown for each of the four types of similarity scores corresponding to the four states of one putative neuron. The y-axis of each plot represents the probability density (the proportion of observations in each bin divided by the bin width). The x-axis indicates similarity score: correlation coefficient for X-corr, waveform, and autocorr; change in log mean rate for the bottom panel.
	The black area in the plot indicates the similarity scores from the same putative neuron along the timeline, which were classified as belonging to the same neuron by the algorithm.
	The Kolmogorov-Smirnov (KS) test results between adjacent stages indicate that, except for the waveform metric, the cross-correlation, autocorrelation, and mean rate metrics do not provide significant evidence to reject the hypothesis of belonging to the same distribution.
}
\label{fig_rslt_real_analysis} 
\end{figure*}

\subsection{Neuron deformation}
\label{dis_deform}
In this study, our focus is specifically on continuous and gradual changes, which are commonly caused by shifts or rotations in the position of cells due to pressure fluctuations near the probe or the effects of pharmacological drugs.
Since these changes typically manifest within a small magnitude, even a fixed classification algorithm can yield satisfactory results in short-duration datasets.
When the time span extends and cumulative errors accumulate, the tracking of neurons can be compromised due to significant deformations. This presents a challenge in the long-term classification of spikes.

We selected an inferred neuron from the classification results of numerous channels in dataset\_real2, which exhibited noticeable waveform deformations. 
The average spike waveforms for four stage are shown in the first row of Figure~\ref{fig_rslt_real_analysis}. 
Drawing insights from \cite{fraser2012recording}, we employed a variety of validation metrics to evaluate the consistency of the inferred neuron.
We characterized the distribution of similarity scores for the inferred neuron at different stages from four perspectives: cross-correlation, waveform, autocorrelation, and mean rate.
The Kolmogorov-Smirnov (KS) tests on the score distributions of adjacent stages suggest a strong likelihood that the neurons represented by the waveforms in these four stages belong to the same entity, which is consistent with our results.

We further evaluated this segment using other state-of-the-art methods again. 
The findings demonstrate that classifiers such as HerdingSpikes2, Mountainsort4, and Tridesclous tend to classify the distorted waveforms in the channels as multiple neurons. 
In contrast, Kilosort and IronClust can only accurately identify a subset of the distorted waveforms, demonstrating inferior performance compared to our method. 
Analysis of the classification results leads us to speculate that Kilosort's template update frequency and IronClust's chunk conjunction miss may contribute to this discrepancy.

Notably, during manual spike sorting, neurons exhibiting considerable drift were erroneously assigned to multiple units, thereby introducing bias into the ground truth. 
Consequently, this issue may potentially influence subsequent analyses and interpretations in the field.

\subsection{End-to-end solutions}
End-to-end solutions offer a promising approach for addressing the spike sorting problem by directly transforming raw or filtered traces into the spike trains of the different spike sorted neurons.
Several solutions have been proposed in different scenarios.
Kilosort\cite{Pachitariu2016} periodically updates templates using a generative model. 
It applies spatial masking and utilizes SVD of spatiotemporal waveforms. 
This makes it particularly well-suited for online applications involving high-density electrode arrays.
Ironclust\cite{jun2017fully} theoretically rejects slow changes by computing time blocks to achieve drift resistance. 
Therefore, it can be considered to have adaptive capabilities as well.

Our method is specifically designed for signal classification using a single channel, without utilizing channel information. 
As a result, our approach can be deployed simultaneously on all channels. 
However, incorporating inter-channel information can offer significant benefits in terms of understanding spatial characteristics, improving classification accuracy, and facilitating the clustering of the same neuron across multiple channels.
Nevertheless, the practical implementation of inter-channel approaches can be labor-intensive and time-consuming, primarily due to the diverse probes atlas. 
The variability in probe configurations presents a challenge to the generalizability of different end-to-end methods.

\subsection{Advantages of neuromorphic design}
A potential advantage of the NeuSort approach is the ultra-low computational cost of implementing neuromorphic chips. 
This is why we preprocess the input signal into spike trains before further processing.
With the increasing demand for large-scale neural recordings, transmitting such a vast amount of data through limited bandwidth has become a recent bottleneck. 
By performing spike sorting prior to signal transmission, where only the sorted spikes are transmitted, the transmission cost can be significantly reduced.
However, it is important to note that neural cells are highly sensitive to temperature. 
Even a slight temperature increase (less than $0.5^\circ$C) can lead to cell degeneration, potentially irreversible, posing challenges to the power consumption of hardware.
To address this issue, the spike sorting process must be computationally efficient for implementation on an implantable device. 
The inherent power efficiency of neuromorphic architectures provides a natural advantage in this regard.
The asynchronous timing characteristic of neuromorphic architectures allows for energy consumption only during spike operations. 
Additionally, the collocated processing and memory design of neuromorphic chips, in contrast to the separate computation and memory organization of traditional Von Neumann architectures, enhances efficiency and reduces space requirements.
Therefore, the NeuSort approach, leveraging neuromorphic chips, offers an ideal solution for the realization of intracranial Brain-Computer Interface (BCI) devices.

Currently, several studies have explored hardware implementations of spike sorting \cite{Werner2016, mukhopadhyay2021power}.
Considering the complexity and power consumption constraints in hardware, these approaches often rely on template matching rather than deep learning methods. 
Moreover, they typically utilize a restricted set of features, such as frequency and peak-to-peak amplitude, for classification purposes.
To the best of our knowledge, our proposed method, NeuSort, is the first neuromorphic model that can be applied online with plasticity updating mechanisms.

\subsection{Limitations and future work}
One of the limitations is that the recognition rate of our method is fixed.
It can be further optimized by refining the learning rate settings. 
Instead of utilizing a fixed learning rate, we can employ a data-driven variable learning rate approach that adapts to different datasets. 
By doing so, we can enhance the adaptability and performance of our method across various experimental scenarios.

\section{Conclusion}
\label{section_conclusion}
In this paper, we propose NeuSort, a novel automatic adaptive spike sorting approach based on neuromorphic models.
With the characteristics of neuromorphic models, NeuSort can learn to sort spike waveforms in an unsupervised manner and adaptively adjust the spike sorter to handle new incoming neurons and waveform deformations.
Experimental results with real data demonstrate that NeuSort achieves state-of-the-art performance while requiring low computational costs.
These results strongly support the potential of employing a neuromorphic model-based spike sorter for ultra-low-cost spike sorting in neuromorphic chips used in implantable BCI devices.
This study presents promising results in applying neuromorphic model-based approaches to spike sorting.

\ack
This work was partly supported by grants from the STI 2030 Major Projects (2021ZD0200400), the Natural Science Foundation of China (61925603, U1909202), the Key Research and Development Program of Zhejiang Province (2023C03001), and the Lingang Laboratory (LG-QS-202202-04).

\section*{Reference}
\bibliography{ref.bib}
\bibliographystyle{IEEEtran}

\end{document}


\title{Appendix for \\
NeuSort: Fully Automated Online Spike Sorting with Neuromorphic Models}

\maketitle

\section{Performance on the synthetic datasets}
Three synthetic datasets are utilized to measure the performance of NeuSort with different conditions: 1) stable spike waveforms, 2) new coming neurons, and 3) spike waveform changes in time. 

Similar to the real dataset experiments, all the spike sorters are evaluated in a pseudo-online manner.
A training set containing spike candidates is used for parameter calibration for PCA-KM. 

\begin{figure*}[htbp]
\centering
	\includegraphics[width=\textwidth]{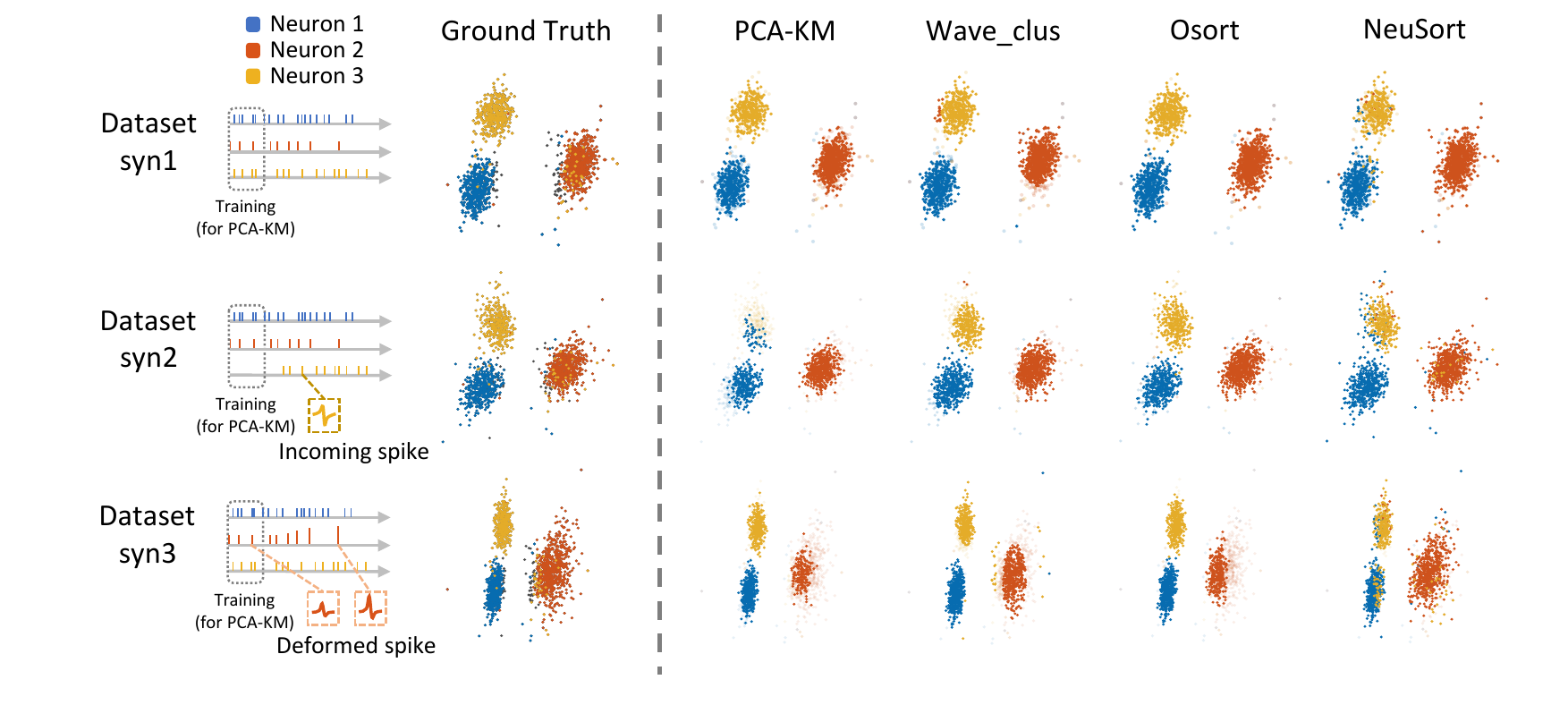}
	\caption{Spike sorting performance on the synthetic datasets.}
\label{fig_rslt_cmp_syn} 
\end{figure*}

The results are presented in Figure~\ref{fig_rslt_cmp_syn} and Table~\ref{tab_method_cmp_on_simulation}. 
In Figure~\ref{fig_rslt_cmp_syn}, the first column sketches the condition of three different synthetic datasets. The second column specifies the ground truth label for spike sorting, with one color representing one neuron (blue as 1, red as 2, yellow as 3). 
For visualization convenience, the spike waveforms are decomposed by PCA, and the first two components are plotted. Rest columns present the spike sorting results of different spike sorters. 
In each subfigure, dots with bright colors refer to a correct detection, while those with light colors are incorrect or missed detection.

\subsection{Synthetic datasets description}
We simulate different conditions of neural activities in the synthetic dataset to evaluate the performance of NeuSort. Specifically, we utilize three situations: 
Stable firing neurons, new coming neurons, and neurons with deformed waveforms.

\begin{itemize}
	\item \textbf{dataset\_syn1.}
		Synthetic dataset 1 (\textit{dataset\_syn1}) is constructed to test the performance of NeuSort in normal online situations.
		Following the methods in \cite{Chaure2018,Quiroga2004}, the waveforms are simulated with three neurons that fire at stable firing rates ranging from 1 to 10 Hz with a renewal Poisson process with a refractory period of 3 ms. 
		For each neuron, one predefined and well-separated waveform is used and rescaled maximum value to $[-1, 1]$.
	\item \textbf{dataset\_syn2.}
		Synthetic dataset 2 (\textit{dataset\_syn2}) is designed to evaluate the performance with new coming neurons.
		One neuron is chosen to be inhibited from firing for a period in the beginning, i.e., only two neurons fire during the training phase for method PCA-KM. 
	\item \textbf{dataset\_syn3.}
		Synthetic dataset 3 (\textit{dataset\_syn3}) contains a neuron whose waveform deforms slightly over time.
		According to the findings presented in \cite{kumar2022tracking}, we conclude that some waveform deformations can be simplified as global scaling.
		To maintain a reasonable limit on the extent of deformation, we impose a restriction of 2 on the maximum deformation ratio of the neuron, meaning that the waveform's amplitude cannot exceed twice its initial value.
		Although the observed neural waveforms do not exhibit such a large scaling behavior during actual acquisition, we enforce this property for improved evaluations.
\end{itemize}

\begin{table*}
\centering
\caption{Results on the synthetic datasets.}
\label{tab_method_cmp_on_simulation}
\begin{threeparttable}
	\begin{tabular}{ll cccc}
	\toprule
	\multicolumn{1}{l}{} & \multicolumn{1}{l}{} & \multicolumn{4}{c}{\textbf{Method}} \\
	\cmidrule(rl){3-6}
	\textbf{Dataset} & \textbf{Criterion} & \textbf{PCA-KM} & \textbf{Wave\_clus} & \textbf{Osort} & \textbf{NeuSort} \\
	\midrule
	\multirow{4}{*}{\textit{dataset\_syn1}}
	& Valid neurons & 3 & 3 & 3 & 3 \\
	& Accuracy & 0.93 & \textbf{0.96} & 0.95 & 0.95 \\
	& Recall & 0.96 & \textbf{0.99} & 0.97 & 0.96 \\
	& Precision & 0.97 & 0.97 & \textbf{0.98} & 0.98 \\
	\midrule
	\multirow{4}{*}{\textit{dataset\_syn2}}
	& Valid neurons & 2 & 3 & 3 & 3 \\
	& Accuracy & 0.60 & 0.89 & \textbf{0.95} & 0.94 \\
	& Recall & 0.72 & 0.94 & \textbf{0.98} & 0.97 \\
	& Precision & 0.79 & 0.94 & \textbf{0.97} & 0.97 \\
	\midrule
	\multirow{4}{*}{\textit{dataset\_syn3}}
	& Valid neurons & 3 & 3 & 3 & 3 \\
	& Accuracy & 0.71 & 0.78 & 0.79 & \textbf{0.85} \\
	& Recall & 0.74 & 0.82 & 0.84 & \textbf{0.99} \\
	& Precision & 0.94 & 0.94 & 0.94 & \textbf{0.94} \\
	\bottomrule
	\end{tabular}
\end{threeparttable}
\end{table*}

\subsection{Performance with stable waveforms}
With \textit{dataset\_syn1}, where the spike waveforms are primarily stable, all the spike sorters achieve high performances. It can be seen from the ground truth that all the detected spikes are correctly stained, and only a few spikes far away from their cluster center are not detected or incorrectly stained.
Specifically, the accuracies are 0.93, 0.96, 0.95, and 0.95 for PCA-KM, Wave\_clus, Osort and NeuSort respectively; the recalls are 0.96, 0.99, 0.97 and 0.96 respectively; and the precisions are 0.97, 0.97, 0.98, and 0.98 respectively. 
Results show that all sorters can distinguish three different neurons, and NeuSort shows high statistical performance for online spike sorting.

\subsection{Performance with new coming neurons}
We evaluate the performance with new coming neurons with \textit{dataset\_syn2}. As shown in Figure~\ref{fig_rslt_cmp_syn}, neuron 3 occurs later in the neural signals. 
The PCA-KM method, where the parameters are fixed after training, fails to detect neuron 3 since the neurons are not presented in the training phase. 
Thus, although PCA-KM works well in an offline manner, the online performance degrades due to minimal online adaptation ability, and only two valid neurons are correctly detected. 
The new coming neuron is correctly detected with Wave\_clus, Osort, and NeuSort. 
Specifically, the recalls are 0.72, 0.94, 0.98, and 0.97 for PCA-KM, Wave\_clus, Osort, and NeuSort, respectively; and the precisions are 0.79, 0.94, 0.97 and 0.97 for PCA-KM, Wave\_clus, Osort, and NeuSort, respectively. 
Results show that NeuSort can cope with new coming neurons in online spike sorting tasks.

\subsection{Performance with waveform deformations}
In online experiments, probe drift or cell distortion can cause slight changes in the spike waveforms of a neuron. As hinted in Figure~\ref{fig_rslt_cmp_syn} (the third row), the waveform of neuron 2 changes in time. 
Usually, the waveform deformation causes a performance decrease due to fixed decoders. 
All four sorters capture three valid neurons. With the PCA-KM method, only 0.54 of the spikes of neuron 2 are detected, and the missing detections are mainly after the waveform deformation. 
Thus, PCA-KM obtains a low accuracy of 0.71 with \textit{dataset\_syn3}. With Wave\_clus and Osort, the waveform deformation is difficult to capture, where only 0.70 and 0.44 of the spikes are detected for neuron 2. NeuSort achieves the best performance under this condition, tracing 0.89 of the spikes of neuron 2. 
The accuracy of NeuSort is 0.85, which is 0.07 and 0.06 higher than Wave\_clus and Osort, respectively. 
Results demonstrate that NeuSort can effectively cope with waveform deformations to improve the robustness of online spike sorting.

\section{Dataset used in analysis of the learning process}

\subsection{Analysis dataset description}
In this experimental section, we generated a new simulated dataset using the same methodology employed for generating dataset\_syn1.
From an existing waveform library, we selected two distinct spike waveforms.
Each waveform was replicated 1000 times, and they were randomly interleaved, resulting in a total of 2000 spike events.
One waveform remained unaltered throughout the entire experiment, while the other waveform underwent a global stretching, doubling its original amplitude after 500 occurrences.
Due to the random occurrence of spikes, the onset of waveform deformation occurred approximately around the 1000th occurrence (specifically, at 1012 occurrences in this experiment).

\subsection{Hybrid dataset description}
To establish a testing environment that is both more challenging and reflective of real-world conditions, we crafted hybrid datasets by amalgamating genuine deformed spike waveforms with diverse levels of noise. 
The authentic spike waveforms, totaling 849 samples, were extracted from dataset\_real1. 
The noise components, procured from publicly accessible datasets, were deliberately chosen to exhibit notable dissimilarity from the spike waveforms under scrutiny. 
These noise elements were then superimposed onto the waveforms at randomized intervals and amplitudes. 
Drawing insights from established methodologies outlined in \cite{Chaure2018,Quiroga2004}, the magnitudes of noise were gauged in relation to their standard deviations, corresponding to proportions of 0.05, 0.1, 0.2, and 0.4 relative to the amplitude of the spike classes. 
Noise levels of 0.05 and 0.1 were designated as readily distinguishable, while levels of 0.2 and 0.4 were classified as posing greater difficulty in differentiation. 
Following the generation of background noise, the spike waveforms were systematically and randomly inserted into the noise signal.

\begin{table*}[ht]
    \centering
	\caption{Results on hybrid datasets.}
	\label{tab_rslt_hybrid}
	\resizebox{\linewidth}{!}{
	\begin{threeparttable}
    \begin{tabular}{l|l|l|c|ccc|cccc}
		\toprule
        \makecell{Noise \\ level} & \makecell{\# Spike \\ detected} & Method & \makecell{\# Spike \\ assigned} & \# tp & \# fn & \# fp & accuracy & precision & recall & F-score \\ 
		\toprule
		0.05 & 823 & PCA-KM & 560 & 512 & 263 & 48 & 0.91 & 0.66 & 0.91 & 0.77 \\ 
        ~ & ~ & Wave\_clus & 794 & 712 & 29 & 82 & 0.90 & 0.96 & 0.90 & 0.93 \\ 
        ~ & ~ & Osort & 690 & 682 & 133 & 8 & 0.99 & 0.84 & 0.99 & 0.91 \\ 
        ~ & ~ & NeuSort & 781 & 702 & 42 & 79 & 0.90 & 0.94 & 0.90 & 0.92 \\ 
		\midrule
        0.1 & 777 & PCA-KM & 547 & 482 & 230 & 65 & 0.88 & 0.68 & 0.88 & 0.77 \\ 
        ~ & ~ & Wave\_clus & 727 & 563 & 50 & 164 & 0.77 & 0.92 & 0.77 & 0.84 \\ 
        ~ & ~ & Osort & 645 & 519 & 132 & 126 & 0.80 & 0.80 & 0.80 & 0.80 \\ 
        ~ & ~ & NeuSort (our) & 727 & 662 & 50 & 65 & 0.91 & 0.93 & 0.91 & 0.92 \\ 
		\midrule
        0.2 & 701 & PCA-KM & 481 & 274 & 220 & 207 & 0.57 & 0.55 & 0.57 & 0.56 \\ 
        ~ & ~ & Wave\_clus & 602 & 328 & 99 & 274 & 0.54 & 0.77 & 0.54 & 0.64 \\ 
        ~ & ~ & Osort & 599 & 405 & 102 & 194 & 0.68 & 0.80 & 0.68 & 0.73 \\ 
        ~ & ~ & NeuSort (our) & 626 & 571 & 75 & 55 & 0.91 & 0.88 & 0.91 & 0.90 \\ 
		\midrule
        0.4 & 541 & PCA-KM & 308 & 123 & 233 & 185 & 0.40 & 0.35 & 0.40 & 0.37 \\ 
        ~ & ~ & Wave\_clus & 472 & 260 & 69 & 212 & 0.55 & 0.79 & 0.55 & 0.65 \\ 
        ~ & ~ & Osort & 424 & 228 & 117 & 196 & 0.54 & 0.66 & 0.54 & 0.59 \\ 
        ~ & ~ & NeuSort (our) & 477 & 428 & 64 & 49 & 0.90 & 0.87 & 0.90 & 0.88 \\ 
		\bottomrule
    \end{tabular}
	\end{threeparttable}
} 
\end{table*}

\subsection{Performance on hybrid dataset}
Table~\ref{tab_rslt_hybrid} presents a comprehensive overview of the statistical outcomes obtained from the application of NeuSort in comparison with alternative methods, utilizing the hybrid datasets.
In the initial two sections characterized by lower noise levels, a noteworthy trend emerges: all methods, with the exception of PCA-KM, showcase commendable recognition capabilities. Specifically, the F-scores for Wave\_clus, Osort, and NeuSort stand at 0.93, 0.91, and 0.92, respectively when noise level is 0.05.
However, as the noise levels progressively increase to higher tiers, a discernible pattern emerges whereby the discernment rates of all comparative methods undergo a discernable reduction. 
For instance, when noise level reaches 0.4, the F-scores for PCA-KM, Wave\_clus, and Osort register values of 0.37, 0.65, and 0.59, respectively.
Notably, our proposed approach upholds a robust F-score across all four distinct noise levels, manifesting a more gradual decline in identification rates when juxtaposed with the comparative techniques. 
This observation highlights the heightened resilience of our method in challenging noise environments.

\bibliography{./ref.bib}
\bibliographystyle{IEEEtran}

\vfill